\documentclass[letterpaper,10pt,conference]{ieeeconf}
\IEEEoverridecommandlockouts
\usepackage[T1]{fontenc}
\usepackage[utf8]{inputenc}
\usepackage{amsmath,amssymb,mathrsfs}
\usepackage{graphicx}
\usepackage{booktabs,array,tabularx,multirow}
\usepackage{cite}
\usepackage{url}
\usepackage{siunitx} 
\usepackage{stfloats}
\graphicspath{{figures/}}
\newcommand{\tabnote}[1]{\par\smallskip\begin{minipage}{\linewidth}\footnotesize #1\end{minipage}}

\title{FloCOS: Predicting Compact Forward Occupancy Sets of Surrounding Vehicles via Conditional Flow Matching}

\author{Chuansheng Dong$^{1}$, Yinxiao Zhan$^{2}$, Zongdian Li$^{3}$, Peng Hou$^{1}$, Jin Wang$^{1,\ast}$, and Jiayu Fan$^{1,\ast}$%
\thanks{$\ast$Corresponding authors: wjin1985@suda.edu.cn; jyfan@suda.edu.cn.}%
\thanks{$^{1}$Chuansheng Dong, Peng Hou, Jin Wang, and Jiayu Fan are with the School of Future Science and Engineering, Soochow University, Suzhou 215222, China, {\tt\small csdong@stu.suda.edu.cn, penghou@suda.edu.cn, wjin1985@suda.edu.cn, jyfan@suda.edu.cn}.}%
\thanks{$^{2}$Yinxiao Zhan is with the College of Control Science and Engineering, Zhejiang University, Hangzhou 310027, China, {\tt\small zhanyinxiao@zju.edu.cn}.}%
\thanks{$^{3}$Zongdian Li is with the State Key Laboratory of Blockchain and Data Security, Zhejiang University, Hangzhou 310007, China, {\tt\small zongdianli@zju.edu.cn}.}%
\thanks{This work was supported in part by the Science and Technology Program of Jiangsu Province (BZ2024062), the Natural Science Foundation of the Jiangsu Higher Education Institutions of China (22KJA520007), Suzhou Planning Project of Science and Technology (2023ss03, SYG2025129, SNG2025010) and Key Laboratory of General Artificial Intelligence and Large Models in Provincial Universities, Soochow University.}%
}

\begin{document}
\maketitle

\thispagestyle{empty}
\pagestyle{empty}

\begin{figure*}[b]
\centering
\includegraphics[width=\textwidth]{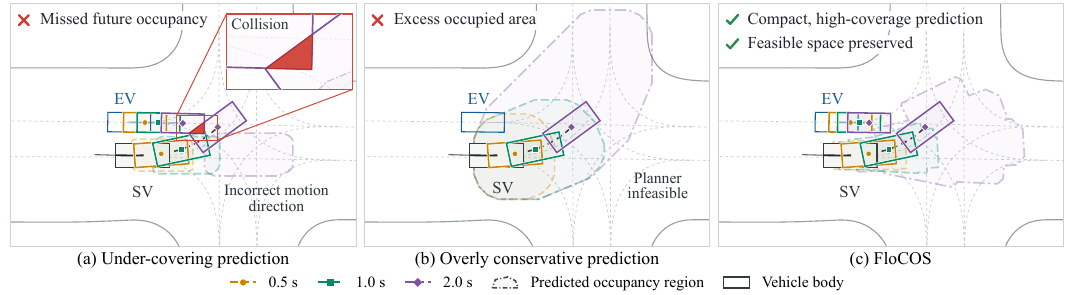}
\caption{Effects of occupancy prediction on planning. All three panels use the same observation history and ego state. Left: undercoverage leaves part of the surrounding vehicle's ground-truth occupancy uncovered, exposing the ego plan to collision risk. Middle: an overly conservative forward occupancy set blocks an otherwise feasible passage. Right: FloCOS covers the ground-truth occupancy in this scene and preserves space for the ego vehicle to pass.}
\label{fig:motivation}
\end{figure*}

\begin{figure*}[t]
\centering
\includegraphics[width=\textwidth]{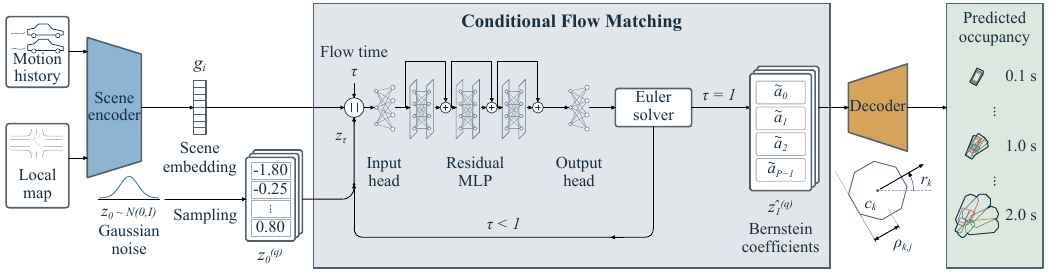}
\caption{FloCOS generates temporal geometry coefficients from vehicle histories and map features and decodes them into sequences of convex components. The rightmost column shows the union of all eight components at selected physical prediction times; intermediate nodes and some individual components are omitted for clarity. Here, flow time $\tau\in[0,1]$ parameterizes the ODE integration path.}
\label{fig:architecture}
\end{figure*}

\begin{abstract}
Existing methods for predicting surrounding vehicles' forward occupancy may under-approximate the ground-truth occupancy, exposing the ego vehicle to collision risk, or produce overly conservative regions that unnecessarily restrict feasible space and cause planning failures. To address these issues, we propose FloCOS, a conditional flow-matching framework that predicts compact forward occupancy sets with high occupancy coverage. FloCOS represents the forward occupancy set as a time-varying union of half-space convex polygons parameterized by temporal basis coefficients. Conditional flow matching captures the conditional distribution of these coefficients to govern the temporal evolution of the convex components, trained under coverage and union-area losses that penalize missed occupancy and excess area. For downstream trajectory optimization, the predicted nonconvex union is directly incorporated into hyperplane-based collision-avoidance constraints. Evaluations on the Waymo Open Motion Dataset show that FloCOS generates compact forward occupancy sets with high coverage and achieves a higher interaction completion rate than evaluated baselines. Real-world vehicle experiments further confirm its ability to support collision-free planning in interactive scenarios.
\end{abstract}

\section{Introduction}
\label{sec:introduction}
Motion prediction is a key component of autonomous driving because the ego vehicle must anticipate the future behavior of surrounding traffic participants before planning safe and efficient maneuvers~\cite{shi2022motion,jiang2023motiondiffuser,yan2025trajflow}. For optimization-based motion planning, these predictions ultimately determine the spatial regions excluded by collision-avoidance constraints~\cite{zhang2020optimization,guthrie2022differentiable}. Uncertainty in collaborative perception~\cite{xu2025codyntrust} and real-to-sim domain shifts~\cite{zuo2025ralad} pose challenges to reliable autonomous driving. Because future behavior depends on uncertain driving intent and traffic interaction, the predicted regions must account for multiple plausible behaviors. If a predicted region fails to cover the ground-truth occupancy, the planner may treat space that the vehicle can actually occupy as collision-free. Conversely, overly enlarged regions unnecessarily shrink the feasible space and may render planning infeasible in confined environments~\cite{shao2024dynamic,zhou2025robust}, as illustrated in Fig.~\ref{fig:motivation}. A practical prediction representation for motion planning must simultaneously achieve high coverage of plausible future occupancy and geometric compactness.

Modern motion predictors can represent complex future behavior, but their native outputs are commonly trajectories or probability distributions instead of bounded geometric sets that can be enforced directly in collision-avoidance constraints. Converting these predictions into forward occupancy sets introduces an additional design problem: vehicle geometry and prediction uncertainty must be represented without excluding unnecessary free space. This problem becomes particularly acute when plausible future occupancies are spatially separated, because a single convex enclosure also occupies the space between them. Representing forward occupancy as a union of compact convex components can naturally capture separated regions without filling the intervening free space.
To this end, we propose FloCOS, which predicts compact forward occupancy sets of surrounding vehicles as unions of occupancy components. Conditional flow matching~\cite{lipman2022flow} generates temporally coherent component sequences from scene context, and the ground-truth occupancy at each prediction step supervises the geometry passed to the planner. The training objectives jointly penalize missed occupancy coverage and unnecessary occupied area over the occupancy union.

Our main contributions are as follows:
\begin{enumerate}
\item We propose FloCOS to predict compact forward occupancy sets of surrounding vehicles by a conditional flow-matching method. This method achieves sufficient and tight coverage of possible forward occupancy regions of surrounding vehicles.
\item We integrate the predicted occupancy unions into trajectory optimization through hyperplane-based collision avoidance constraints. This allows the planner to use the nonconvex occupancy representation without replacing it with an approximate convex enclosure.
\item We validate FloCOS on the Waymo Open Motion Dataset and in real-world vehicle experiments. The results demonstrate that FloCOS produces more compact occupancy predictions than existing methods and enables collision-free planning for ego vehicles.
\end{enumerate}

\section{Related Works}
\label{sec:related}
\textit{Motion prediction for autonomous driving:}
Recent predictors largely rely on transformer and generative architectures to capture multimodal future behaviors. For instance, MTR uses learned motion queries to distinguish motion modes, MotionDiffuser models joint multi-agent trajectory distributions with diffusion, and TrajFlow applies flow matching for efficient multi-trajectory sampling~\cite{shi2022motion,jiang2023motiondiffuser,yan2025trajflow}. Beyond pure forecasting, several studies align prediction with downstream control: TIP incorporates downstream task utility, and DIPP couples prediction with a differentiable planner~\cite{9982100,huang2023differentiable}. Another direction replaces discrete trajectories with spatial grids, as seen in Occupancy Flow Fields and ImplicitO, or unifies occupancy forecasting with planning objectives as in P3 and MP3~\cite{mahjourian2022occupancy,agro2023implicit,sadat2020perceive,casas2021mp3}.

\textit{Set-based prediction for motion planning:}
Instead of outputting trajectories or dense grids, set-based predictors construct bounded forward-occupancy regions through learned geometry, reachability analysis, or statistical calibration. DOS learns elliptical occupancy sets by balancing future-position coverage against occupied area, which DOSP extends by incorporating vehicle pose and dimensions for planning~\cite{shao2024towards,shao2024dynamic}. Data-driven approaches also directly regress convex polygonal or zonotopic representations~\cite{xiang2024learning,shamsah2024socially}. Alternatively, reachability-based methods derive forward occupancy sets from vehicle dynamics, traffic rules, or learned control uncertainty~\cite{koschi2020set,zhou2025robust}.

A complementary line of work calibrates or extracts compact prediction regions from existing predictors: MMCPR optimizes convex shape templates~\cite{tumu2024multi}, FORCE-OPT derives calibrated reachable regions from trajectory-prediction distributions~\cite{chakraborty2026safety}, and conformal-prediction-based planners use statistically valid regions for safety and recursive-feasibility reasoning~\cite{lindemann2023safe,stamouli2024recursively}. ZAPP further converts discrete probabilistic predictions into zonotopic representations for continuous-time collision avoidance~\cite{paparusso2024zapp}. FloCOS differs by learning a union of compact occupancy components directly from ground-truth occupancy supervision and optimizing the geometry used by the planner during training.

\section{Problem Formulation}
\label{sec:problem}
Consider an ego vehicle with state $\xi$, control $u$, known dynamics, and convex occupancy $B(\xi)$, navigating a workspace $W\subset\mathbb R^2$ from an initial state $\xi_{\mathrm{init}}$ toward a goal state $\xi_{\mathrm{final}}$. Let $\mathcal I$ denote the set of surrounding vehicles whose forward occupancy sets are predicted for ego-vehicle planning. At prediction times $t_k=k\Delta t$, $k=1,\ldots,N$, the ground-truth occupancy of each vehicle $i\in\mathcal I$ is unknown to the planner; FloCOS predicts forward occupancy sets at these times to construct collision-avoidance constraints for the ego vehicle.

For each surrounding vehicle $i\in\mathcal I$, treated as the prediction target, let $\mathbf X_i^-$ collect the histories and attributes of all scene vehicles expressed in its current local frame, and let $M_i$ denote the corresponding local map. The superscript $-$ denotes past and current vehicle states only. Let $O_{i,k}^\star\subset\mathbb R^2$ denote the ground-truth occupancy of vehicle $i$ at prediction step $k$. Given its center $p_{i,k}$, heading $\psi_{i,k}$, length $l_{i,k}$, and width $w_{i,k}$,
\begin{equation}\label{eq:occupancy}
O_{i,k}^\star=p_{i,k}+R(\psi_{i,k})\Bigl(
\bigl[-\tfrac{l_{i,k}}2,\tfrac{l_{i,k}}2\bigr]
\times\bigl[-\tfrac{w_{i,k}}2,\tfrac{w_{i,k}}2\bigr]\Bigr),
\end{equation}
where $R(\psi_{i,k})$ is the planar rotation matrix.

The predictor outputs $Q$ sequences of convex occupancy components:
\begin{equation}\label{eq:union}
\begin{aligned}
\widehat O_{i,k}^{(q)}&=\{s\in\mathbb R^2:U_{i,k}^{(q)}s\preceq h_{i,k}^{(q)}\},\\
\widehat O_{i,k}&=\bigcup_{q=1}^{Q}\widehat O_{i,k}^{(q)},
\end{aligned}
\end{equation}
where $U\in\mathbb R^{J\times2}$ and $h\in\mathbb R^J$ define $J$ half-spaces. Each occupancy component $\widehat O_{i,k}^{(q)}$ is nonempty and bounded, and their union $\widehat O_{i,k}$ can be nonconvex and disconnected. Here, index $q$ simply denotes the $q$-th generated component sequence. Surrounding vehicles are predicted independently without cross-vehicle mode coupling.

We say that the forward occupancy of vehicle $i$ is simultaneously covered over the first $H\leq N$ prediction steps if
$O_{i,k}^\star\subseteq\widehat O_{i,k}$ for every $k=1,\ldots,H$.
At each step, coverage is evaluated against
$\widehat O_{i,k}=\bigcup_{q=1}^{Q}\widehat O_{i,k}^{(q)}$,
where each $q$ indexes a complete time-varying component sequence over the prediction horizon.
If this condition holds and the ego vehicle avoids the predicted occupancy union, collisions with the ground-truth occupancy are excluded at these prediction steps.
We therefore evaluate occupancy coverage, area ratio, and planning outcomes separately.

\section{Method}
\label{sec:method}
FloCOS parameterizes the time-varying geometry of each occupancy-component sequence with Bernstein basis coefficients, models their conditional distribution with flow matching, and decodes the generated coefficients into occupancy unions. Training combines coefficient-space flow matching with losses defined on the decoded sets. Fig.~\ref{fig:architecture} summarizes the generation process.

\subsection{Convex Geometry and Temporal Parameterization}
Omitting vehicle and time indices, choose $J\geq4$ as a multiple of four and define reference normals $\bar n_j=(\cos\theta_j,\sin\theta_j)^\top$, where $\theta_j=2\pi(j-1)/J$ and $j=1,\ldots,J$. Each occupancy component is specified by a center $c$, a rotation $R$, and $J/2$ positive half-widths $\rho_j$. Letting $n_j=R\bar n_j$, the half-space offsets are
\begin{equation}\label{eq:paired}
\begin{aligned}
h_j&=n_j^\top c+\rho_j,\\
h_{j+J/2}&=-n_j^\top c+\rho_j,\qquad j=1,\ldots,J/2.
\end{aligned}
\end{equation}
Each occupancy component is therefore the intersection of the strips $|n_j^\top(s-c)|\leq\rho_j$. Finite positive half-widths ensure nonemptiness because $c$ is feasible, while two orthogonal strip pairs confine the set to a finite rectangle and therefore ensure boundedness. The half-space intersection is convex and centrally symmetric, with at most $J$ edges. These geometric properties alone do not imply coverage of the ground-truth occupancy.

When plausible future occupancies are spatially separated, a single convex set covering all of them must also include the space between them. Multiple convex occupancy components allow the predicted occupancy union to represent these separated regions without necessarily occupying the intervening space, thereby preserving more free space for downstream planning. 

For a ground-truth occupancy, set $c_k^\star=p_k$ and align $R_k^\star$ with the vehicle heading. Its half-width labels are $\rho_{k,j}^\star=(l_k/2)|\cos\theta_j|+(w_k/2)|\sin\theta_j|$. The two orthogonal axial strips recover the oriented vehicle rectangle, and the remaining oblique strips form tight redundant bounds. To avoid angular discontinuities, rotation is represented by the unit direction $r_k^\star=(\cos\psi_k,\sin\psi_k)^\top$. The ground-truth geometric descriptor is
\[
y_k^\star=
\bigl((c_k^\star)^\top,\log(\rho_k^\star)^\top,(r_k^\star)^\top\bigr)^\top
\in\mathbb R^{d_y},
\qquad d_y=4+J/2.
\]
After temporal reconstruction, the two reconstructed direction coordinates are denoted by $\widetilde r_k$ and normalized to unit length before constructing the rotation matrix.

To compactly parameterize its temporal evolution, the ground-truth descriptor sequence $\{y_k^\star\}_{k=1}^{N}$ is represented using $P<N$ Bernstein basis functions $b_m(t)=\binom{P-1}{m}t^m(1-t)^{P-1-m}$, with $\bar t_k=(k-1)/(N-1)$:
\begin{equation}\label{eq:bernstein}
\begin{aligned}
y_k&=\sum_{m=0}^{P-1}b_m(\bar t_k)a_m,\qquad A^\star=\mathsf B^\dagger Y^\star,\\
z_1&=\operatorname{vec}\!\left((A^\star-\mu_A)\oslash\sigma_A\right),
\end{aligned}
\end{equation}
where $a_m\in\mathbb R^{d_y}$ are the Bernstein coefficients, $\mathsf B_{k,m+1}=b_m(\bar t_k)$, $Y^\star$ stacks the occupancy labels as rows, and $\mathsf B^\dagger$ fits the coefficients by least squares. Training-set statistics $\mu_A$ and $\sigma_A$ standardize the coefficients, and $\oslash$ denotes elementwise division. This reduces the generation dimension from $Nd_y$ to $Pd_y$. The decoder $\mathscr D$ reverses the standardization, reconstructs the sequence, recovers positive half-widths and unit directions, and constructs the occupancy components according to~\eqref{eq:paired}.

\subsection{Scene-Conditioned Flow Matching}
The scene encoder produces a target-conditioned representation $g_i=E_\varphi(\mathbf X_i^-,M_i)$. Shared temporal convolutions encode the scene-vehicle histories together with valid-observation masks, observation intervals, class labels, and dimensions. Target-conditioned attention aggregates information across scene vehicles, while a polyline map branch combines target-conditioned attention with multiple map queries. The resulting scene representation conditions the generation of all occupancy components at every integration step.

The residual vector-field network $v_\theta$ models the conditional coefficient distribution using a linear flow-matching path~\cite{lipman2022flow,tong2023improving,liu2022flow}. Given the encoded scene representation $g_i$ and its target coefficient vector $z_1$, we independently sample $z_0\sim\mathcal N(0,I)$ and $\tau\sim\mathcal U[0,1]$, yielding
\begin{equation}\label{eq:fm}
\begin{aligned}
z_\tau&=(1-\tau)z_0+\tau z_1,\\
\mathcal L_{\mathrm{FM}}&=\mathbb E\!\left[
\frac{\|v_\theta(z_\tau,\tau,g_i)-(z_1-z_0)\|_2^2}{Pd_y}\right].
\end{aligned}
\end{equation}
Flow time $\tau\in[0,1]$ is distinct from physical prediction time; each flow state $z_\tau$ represents Bernstein coefficients for the entire future horizon $t_1,\ldots,t_N$. At inference, explicit Euler integration solves
\begin{equation}\label{eq:ode}
\frac{\mathrm dz^{(q)}}{\mathrm d\tau}=v_\theta(z^{(q)},\tau,g_i),
\qquad z^{(q)}(0)=z_0^{(q)},
\end{equation}
and $\mathscr D(z^{(q)}(1))$ yields one occupancy-component sequence. Equation~\eqref{eq:fm} uses freshly sampled Gaussian source variables, whereas set-level training and inference share a fixed source set $\mathcal Z_0=\{\pm\eta_q\}_{q=1}^{Q/2}$, where $Q$ is even and the $\eta_q$ are sampled independently from a standard Gaussian at initialization. Using the same symmetric finite source set aligns the occupancy union optimized during set-level training with the one evaluated at inference and avoids fluctuations due to resampling.

After flow-matching pretraining, the set-level losses are introduced while retaining $\mathcal L_{\mathrm{FM}}$. For probe points $\mathcal P_{i,k}$ on the boundary and in the interior of the ground-truth occupancy, define
\begin{equation}\label{eq:coverage-loss}
\begin{aligned}
d_{i,k}^{\cup}(s)&=\min_q\max_j\bigl((n_{i,k,j}^{(q)})^\top s-h_{i,k,j}^{(q)}\bigr),\\
r_i^{(H)}&=\max_{1\leq k\leq H}\max_{s\in\mathcal P_{i,k}}d_{i,k}^{\cup}(s),\\
\mathcal L_{\mathrm{cov}}^{(H)}&=\operatorname{TailMean}_{\beta}
\bigl(\{[r_i^{(H)}]_+\}_{i\in\mathcal B}\bigr).
\end{aligned}
\end{equation}
By construction, $\max_j\bigl((n_{i,k,j}^{(q)})^\top s-h_{i,k,j}^{(q)}\bigr)\leq0$ if and only if $s\in\widehat O_{i,k}^{(q)}$, so $d_{i,k}^{\cup}(s)\leq0$ indicates $s\in\widehat O_{i,k}$. Taking the maxima over discrete probe points $\mathcal P_{i,k}$ and steps $k\leq H$ provides a tractable piecewise differentiable surrogate for simultaneous containment. The hinge operator $[\cdot]_+$ zeroes out penalties on covered points, and $\operatorname{TailMean}_{\beta}$ averages the worst $\beta$-fraction of sample violations within batch $\mathcal B$.

Coverage alone is insufficient, because enlarging the predicted union can trivially improve coverage while unnecessarily reducing the free space available to the planner. We therefore penalize the normalized union area to encourage compact occupancy predictions.
\begin{equation}\label{eq:area-loss}
\mathcal L_{\mathrm{area}}^{(H)}=\mathbb E_i\!\left[\frac1H\sum_{k=1}^{H}
\frac{\widetilde{\operatorname{Area}}(\widehat O_{i,k})}{\operatorname{Area}(O_{i,k}^\star)}\right].
\end{equation}
During training, the area term is computed numerically. We intersect horizontal and vertical scanlines with each convex occupancy component to obtain intervals. Overlapping intervals are merged, and the resulting union length is integrated by midpoint quadrature. Horizontal and vertical estimates are computed separately within a finite region around the ground-truth center and then averaged.

An occupancy component can expand within regions already covered by other components without increasing union area, so its scale and shape are regularized separately. Let $j_\perp=1+J/4$ and define $\rho_j^\square=\rho_1|\cos\theta_j|+\rho_{j_\perp}|\sin\theta_j|$. Two orthogonal half-widths confine the component to a rectangle of area $4\rho_1\rho_{j_\perp}$, yielding
\begin{equation}\label{eq:regularizers}
\begin{aligned}
\mathcal L_{\mathrm{comp}}&=\mathbb E_{i,q,k}\!\left[
\frac{4\rho_{i,k,1}^{(q)}\rho_{i,k,j_\perp}^{(q)}}{\operatorname{Area}(O_{i,k}^\star)}\right],\\
\mathcal L_{\mathrm{shape}}&=\mathbb E_{i,q,k,j}\!\left[
\left(\frac{\rho_j-\rho_j^\square}{\max(\operatorname{sg}(\rho_j^\square),\delta_\rho)}\right)^2\right].
\end{aligned}
\end{equation}
The shape term omits indices $i,q,k$ for readability; $\operatorname{sg}$ denotes stop-gradient, and $\delta_\rho>0$ keeps the denominator bounded away from zero. This regularizer guides the oblique half-widths toward rectangular proportions without restricting the polygon to four edges.

Given a position scale $s_c>0$, define the geometric descriptor $D_{i,k}^{(q)}=[c_{i,k}^{(q)}/s_c,\log(\rho_{i,k}^{(q)}/\mathrm m),r_{i,k}^{(q)}]$, and let $D_i^{(q)}$ denote the full descriptor sequence. For the ground-truth descriptor $D_i^\star$, we use
\begin{equation}\label{eq:diversity}
\begin{aligned}
\mathcal L_{\mathrm{div}}=\mathbb E_i\Bigg[&\frac1{QN}\sum_{q,k}\|D_{i,k}^{(q)}-D_{i,k}^\star\|_2\\
&-\frac1{2Q^2\sqrt{Nd_y}}\sum_{q,q'}\|\operatorname{vec}(D_i^{(q)}-D_i^{(q')})\|_2\Bigg].
\end{aligned}
\end{equation}
The first term limits deviation from the ground-truth geometry, while the second encourages diversity across generated sequences. Both the coverage and area losses are evaluated at the near horizon $N_h<N$ and at the full horizon $N$, giving the total objective
\begin{equation}\label{eq:total-loss}
\begin{aligned}
\mathcal L={}&\lambda_{\mathrm{FM}}\mathcal L_{\mathrm{FM}}\\
&+\sum_{H\in\{N_h,N\}}\!\left(\lambda_{\mathrm{cov},H}\mathcal L_{\mathrm{cov}}^{(H)}
+\lambda_{\mathrm{area},H}\mathcal L_{\mathrm{area}}^{(H)}\right)\\
&+\lambda_{\mathrm{comp}}\mathcal L_{\mathrm{comp}}+\lambda_{\mathrm{shape}}\mathcal L_{\mathrm{shape}}
+\lambda_{\mathrm{div}}\mathcal L_{\mathrm{div}},
\end{aligned}
\end{equation}
which jointly updates the scene encoder and vector field.

\subsection{Trajectory Optimization Interface}
Let $\mathcal T_i(s)=R_i^{\mathrm p}s+p_i^{\mathrm p}$ denote the rigid transformation from target vehicle $i$'s current local frame to the planner frame, where $R_i^{\mathrm p}$ and $p_i^{\mathrm p}$ are given by its current pose in the planner frame. The corresponding planner-frame occupancy component is $\widetilde O_{i,k}^{(q)}=\mathcal T_i(\widehat O_{i,k}^{(q)})$. Let $\boldsymbol\xi=(\xi_0,\ldots,\xi_N)$ and $\boldsymbol u=(u_0,\ldots,u_{N-1})$, and let $\mathcal F_{\mathrm{ego}}$ collect the initial condition, discrete dynamics, and state, input, and road constraints. A full-horizon hard-constraint formulation is
\begin{equation}\label{eq:planner}
\begin{aligned}
\min_{\boldsymbol\xi,\boldsymbol u}\quad &\Phi(\xi_N,\xi_{\mathrm{final}})
+\sum_{k=0}^{N-1}\Delta t\,\ell(\xi_k,u_k)\\
\mathrm{s.t.}\quad &(\boldsymbol\xi,\boldsymbol u)\in\mathcal F_{\mathrm{ego}},\\
&\operatorname{dist}\!\left(B(\xi_k),\widetilde O_{i,k}^{(q)}\right)\geq d_{\min},\\
&\forall i\in\mathcal I,\ k=1,\ldots,N,\ q=1,\ldots,Q.
\end{aligned}
\end{equation}
where $\ell$ and $\Phi$ are the stage and terminal costs, respectively, and $\operatorname{dist}$ denotes minimum Euclidean set distance. The collision-avoidance constraints are implemented using existing hyperplane-based formulations~\cite{zhang2020optimization,guthrie2022differentiable}. Since both sets already include the full vehicle geometry, $d_{\min}>0$ represents only additional clearance. The occupancy union is enforced componentwise, requiring $|\mathcal I|NQ$ collision-avoidance constraint groups. In the experiments, these constraints are enforced as hard constraints over the near horizon and relaxed over the longer horizon, as detailed in Section~\ref{sec:setup}.

\section{Simulation And Experiments}
\label{sec:experiments}
We first evaluate the prediction quality through simultaneous occupancy coverage and mean area ratio, and then verify downstream planning performance, representation and generation alternatives, and real-world experiments.

\subsection{Experimental Setup}
\label{sec:setup}
\textit{Data and training:}
We use the Waymo Open Motion Dataset (WOMD)~\cite{ettinger2021large} with disjoint training, development, calibration, and test sets, containing 197,464 training windows, 13,984 development windows, and 1000 test scenarios. Each configuration predicts \SI{2}{s} into the future at \SI{0.1}{s} intervals from \SI{1}{s} of history. FloCOS uses $Q=8$, $J=8$, $P=6$, and two Euler integration steps. It is trained with AdamW using a batch size of 4096, a learning rate of $3\times10^{-4}$, weight decay of $10^{-4}$, and a gradient clipping threshold of 5. Training runs for 8,000 steps; the first 200 steps use only the flow-matching loss, after which the set-level losses are introduced.

\textit{Baselines:}
We compare MTR~\cite{shi2022motion}, DenseTNT~\cite{gu2021densetnt}, M2I~\cite{sun2022m2i}, TrajFlow~\cite{yan2025trajflow}, Traffic-Rule Set Prediction~\cite{koschi2020set}, DOS-Ellipse~\cite{shao2024towards}, MTR-GMM + FORCE-OPT~\cite{chakraborty2026safety}, and MTR-Top1 + MMCPR-Joint~\cite{tumu2024multi}. DOS-Ellipse trains an elliptical head on frozen MTR features and center trajectories; FORCE-OPT uses MTR Gaussian-mixture outputs, and MMCPR uses MTR Top-1 residuals. All methods account for vehicle geometry, and their outputs are represented as convex occupancy components, with calibration parameters frozen before testing. All baselines are evaluated using the task-specific adaptations described above under the same evaluation protocol.

\textit{Prediction evaluation:}
For each test scenario, we predict one designated target vehicle at a fixed observation time. We report the empirical simultaneous occupancy coverage rate, defined as the fraction of test cases for which
$O_{i,k}^\star\subseteq\widehat O_{i,k}$ holds for every evaluated prediction step, and the mean area ratio $\operatorname{Area}(\widehat O_{i,k})/\operatorname{Area}(O_{i,k}^\star)$ over scenarios and prediction steps. We evaluate horizons of $H=10$ (\SI{1}{s}) and $H=20$ (\SI{2}{s}). Evaluation uses exact polygon containment and union-area computation.

\textit{Planning evaluation:}
An ego vehicle is inserted into each scenario and executes receding-horizon control, while background vehicles replay their recorded trajectories without reacting to the ego vehicle. All methods use the same dynamics, road constraints, cost function, and termination rules. Collision-avoidance constraints are enforced as hard constraints over the first \SI{1}{s} and as penalized soft constraints over the second \SI{1}{s}. An interaction is counted as completed if the ego vehicle reaches the prescribed passage progress without earlier termination caused by a collision, road violation, or solver failure. Occupancy coverage and interaction completion rate are evaluated separately. Interaction completion rates are reported with 95\% Wilson intervals, which quantify proportion uncertainty for this fixed-composition test set.

\subsection{Illustrated Results}
\subsubsection{How Occupancy Coverage and Compactness Affect Planning}
\begin{figure*}[t]
\centering
\includegraphics[width=\textwidth]{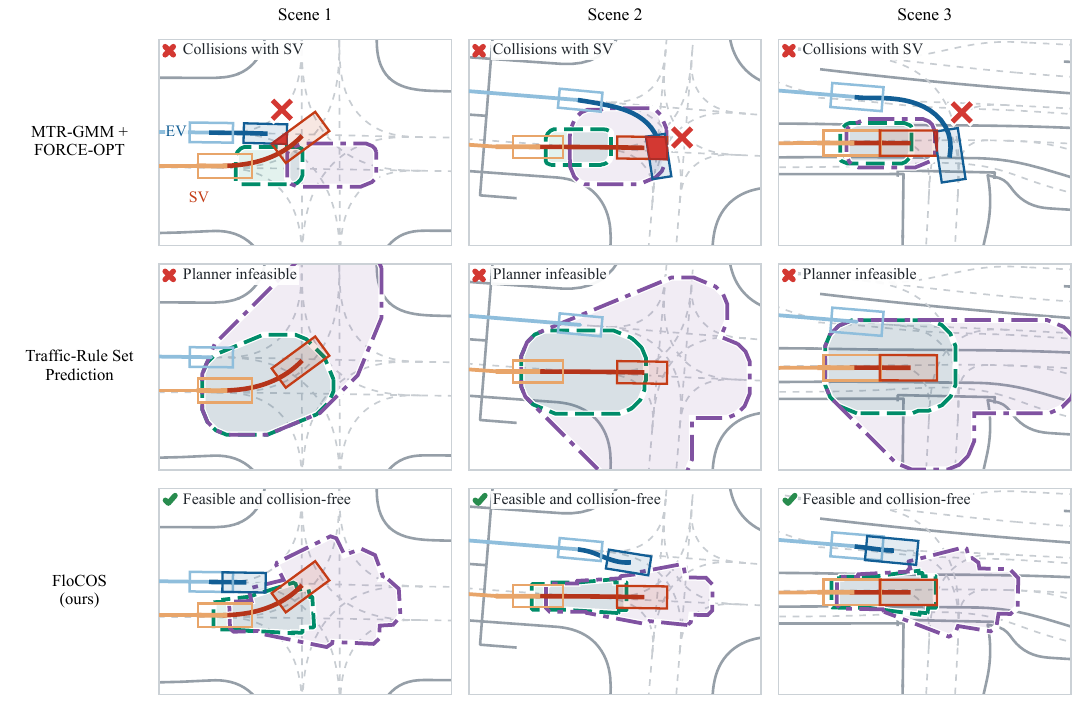}
\caption{Representative planning outcomes in three simulated interactions. Each column shows the same scene and the same planner, with the prediction method varying across rows; rows correspond to MTR-GMM + FORCE-OPT, Traffic-Rule Set Prediction, and FloCOS. Blue and orange--red denote the EV and SV, respectively, with light and dark traces indicating history and future motion. Green and purple dashed boundaries show the predicted SV occupancy at \SI{1}{s} and \SI{2}{s}, respectively.}
\label{fig:prediction-horizon}
\end{figure*}

Fig.~\ref{fig:prediction-horizon} places the three set predictors in the same three simulated interactions. These scenarios directly show the importance of occupancy coverage and compactness to the planner. In the shown interactions, MTR-GMM + FORCE-OPT produces compact but under-covering predictions, causing the planned trajectory to collide with the surrounding vehicle. Traffic-Rule Set Prediction provides sufficient coverage but is overly conservative, leaving the planner without a feasible path. FloCOS predicts compact, high-coverage forward occupancy sets, allowing the planner to generate safe and feasible paths. We therefore evaluate occupancy coverage and compactness jointly, and then assess planning performance under the same planner.

\subsubsection{Occupancy Quality and Planning Performance}
Table~\ref{tab:main} reports prediction and planning results on 1000 independent WOMD test scenarios.

\begin{table*}[t]
\caption{Prediction and Planning Results on 1000 Independent WOMD Test Scenarios}
\label{tab:main}
\centering
\small
\renewcommand{\arraystretch}{1.08}
\begin{tabular*}{\textwidth}{@{\extracolsep{\fill}}cccc@{}}
\toprule
Methods
& Coverage (\%, \SI{1}{s}/\SI{2}{s})
& Area ratio (\SI{1}{s}/\SI{2}{s})
& Completion (\%) [95\% CI]\\
\midrule
MTR~\cite{shi2022motion}
& 88.5 / 88.2
& 2.15 / 3.76
& 47.4 [44.3, 50.5]\\
DenseTNT~\cite{gu2021densetnt}
& 91.8 / 91.4
& 2.48 / 4.88
& 23.0 [20.5, 25.7]\\
M2I~\cite{sun2022m2i}
& 92.8 / 92.5
& 2.58 / 5.16
& 18.9 [16.6, 21.4]\\
TrajFlow~\cite{yan2025trajflow}
& 92.1 / 91.6
& 2.20 / 3.62
& 48.8 [45.7, 51.9]\\
Traffic-Rule Set Prediction~\cite{koschi2020set}
& 86.4 / 83.3
& 5.04 / 8.59
& 3.1 [2.2, 4.4]\\
MTR-GMM + FORCE-OPT~\cite{shi2022motion,chakraborty2026safety}
& 72.8 / 69.7
& 1.61 / 2.18
& 85.2 [82.9, 87.3]\\
MTR-Top1 + MMCPR-Joint~\cite{shi2022motion,tumu2024multi}
& 82.5 / 81.5
& 2.45 / 4.72
& 10.7 [8.9, 12.8]\\
DOS-Ellipse~\cite{shao2024towards}
& 99.6 / 98.8
& 3.02 / 3.98
& 59.1 [56.0, 62.1]\\
\textbf{FloCOS}
& \textbf{99.7 / 99.3}
& \textbf{1.70 / 3.10}
& \textbf{85.6 [83.3, 87.6]}\\
\bottomrule
\end{tabular*}
\tabnote{For coverage and area ratio, values before and after the slash correspond to the \SI{1}{s} and \SI{2}{s} prediction horizons, respectively. Coverage requires containment of the full ground-truth occupancy at every evaluation step. Area ratio is normalized by the ground-truth occupancy area. Completion rates are followed by 95\% Wilson confidence intervals.}
\end{table*}

FloCOS and DOS-Ellipse achieve comparable full-horizon coverage (99.3\% vs. 98.8\%), while FloCOS reduces the area ratio from 3.98 to 3.10 and increases interaction completion from 59.1\% to 85.6\%. FloCOS also achieves higher coverage and lower area than the adapted MTR, DenseTNT, M2I, TrajFlow, and MMCPR-Joint variants. Traffic-Rule Set Prediction illustrates the opposite failure mode, with an area ratio of 8.59 and only 3.1\% completion. FORCE-OPT attains similar completion to FloCOS (85.2\%) but only 69.7\% full-horizon coverage, showing that completion should not substitute for occupancy coverage.

\subsubsection{Representation, Generation, and Set-Level Objectives}
We next ablate key design choices in Table~\ref{tab:ablation}. Traj-CFM retains conditional flow matching but predicts center trajectories prior to calibrated occupancy construction. Deterministic and Gaussian keep our encoder and polygon representation but replace flow matching with multi-head regression and a diagonal Gaussian head, respectively. For the loss ablations, Component-Sum Area replaces the union-area penalty in~\eqref{eq:area-loss} with a scale-matched sum of individual component areas, and Single-Component Coverage forces a single component to cover the entire vehicle footprint in~\eqref{eq:coverage-loss}.

\begin{table*}[t]
\caption{Ablation results for FloCOS design choices.}
\label{tab:ablation}
\centering
\small
\renewcommand{\arraystretch}{1.08}
\begin{tabular*}{\textwidth}{@{\extracolsep{\fill}}cccc@{}}
\toprule
Configuration
& Coverage (\%, \SI{1}{s}/\SI{2}{s})
& Area ratio (\SI{1}{s}/\SI{2}{s})
& Completion (\%)\\
\midrule
\textbf{FloCOS}
& \textbf{99.7 / 99.3}
& \textbf{1.70 / 3.10}
& \textbf{85.6}\\
Traj-CFM
& 99.4 / 66.1
& 2.91 / 3.84
& 9.6\\
Deterministic
& 56.1 / 21.3
& 2.64 / 2.77
& 64.8\\
Gaussian
& 17.7 / 13.4
& 1.73 / 2.99
& 61.1\\
Component-Sum Area
& 26.6 / 26.0
& 2.46 / 4.62
& 57.2\\
Single-Component Coverage
& 99.4 / 99.1
& 2.03 / 3.99
& 78.1\\
\bottomrule
\end{tabular*}
\tabnote{For coverage and area ratio, values before and after the slash correspond to the \SI{1}{s} and \SI{2}{s} horizons, respectively. Component-Sum Area uses a rescaled component-wise area penalty with magnitude matched to the original union-area term. Metrics otherwise follow Table~\ref{tab:main}.}
\end{table*}

Comparing FloCOS with Traj-CFM isolates the effect of the prediction target. Both achieve similar \SI{1}{s} coverage (99.7\% vs. 99.4\%), but FloCOS reduces the \SI{1}{s} mean area ratio by 41.6\% (1.70 vs. 2.91). Traj-CFM's coverage drops to 66.1\% at the \SI{2}{s} prediction horizon. Its overall interaction completion rate is 9.6\%. These results confirm the benefit of directly learning planner-facing set geometry over post-hoc trajectory buffering.

The set-level objective ablations highlight the role of union-aware training. Penalizing the sum of individual component areas (Component-Sum Area) causes coverage to collapse to 26.6\% / 26.0\%, increases the union-area ratio to 2.46 / 4.62, and lowers completion to 57.2\%, showing that shrinking components independently fails to coordinate a tight union. Conversely, Single-Component Coverage maintains high coverage (99.4\% / 99.1\%) but inflates the area ratio to 2.03 / 3.99 and drops completion to 78.1\%, demonstrating the advantage of jointly covering the footprint across multiple convex components.

\subsection{Real-World Experiments}
We further validate FloCOS on two ROSMASTER mobile robots in a head-on encounter and a lateral cut-in scenario. To transfer the WOMD-trained predictor to the small-scale platform, we apply a uniform spatial scale factor that matches the robot footprint area to the median vehicle footprint in WOMD. Positions, dimensions, map geometry, and velocities are scaled prior to network inference (keeping time and heading unscaled), and the predicted sets are inversely scaled back to the physical workspace before planning. FloCOS runs onboard the ego vehicle (EV) in real time. The surrounding vehicle (SV) follows a prescribed path and broadcasts its ground-truth pose so that prediction and planning can be evaluated independently of perception noise.

\begin{figure*}[t]
\centering
\includegraphics[width=\textwidth]{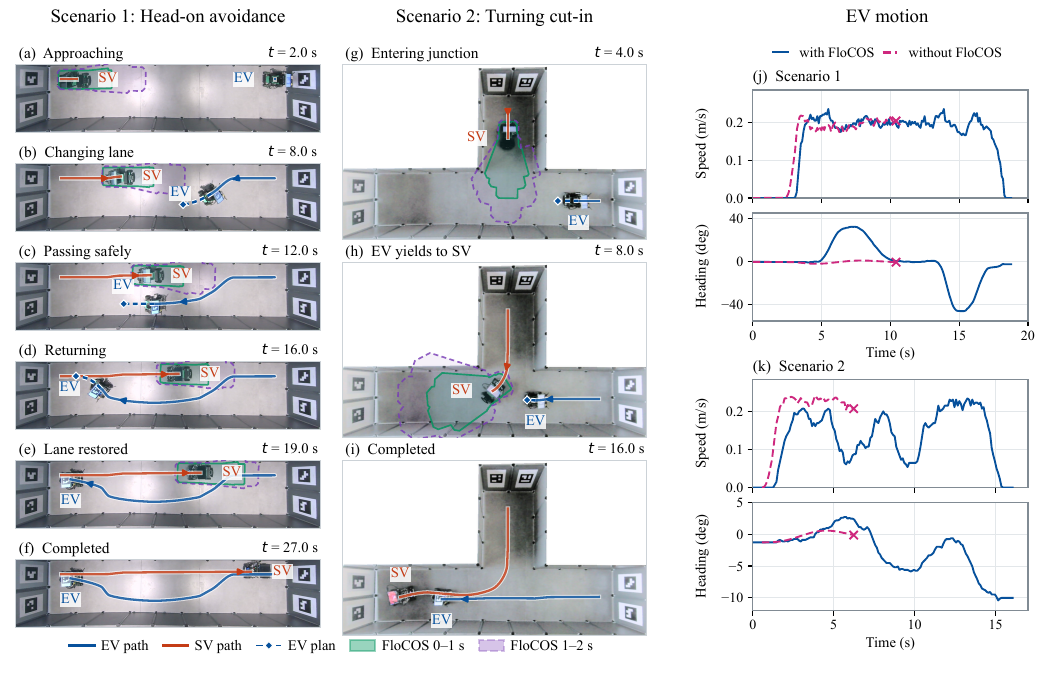}
\caption{Real-world interaction experiments with FloCOS. (a)--(f) Head-on encounter; (g)--(i) lateral cut-in. Panels (j) and (k) respectively present the EV motion data in the head-on and lateral cut-in scenarios. In each panel, the upper plot shows EV speed and the lower plot shows EV heading; solid blue curves correspond to the real-world experiments with FloCOS, and dashed magenta curves show matched simulations without FloCOS. Crosses mark EV--SV collisions in the simulations.}
\label{fig:real-world}
\end{figure*}

In both scenarios, FloCOS successfully enables the planner to solve the motion-planning task. The two interactions elicit different EV responses. In the head-on case, the EV first moves out of the conflict region, passes the SV, and then returns toward its original lane. In the cut-in case, the EV yields before the SV enters the intersection and continues toward the goal after the SV clears it. Without the FloCOS prediction sets, simulations that replay the recorded SV trajectories result in EV--SV collisions in both scenarios, as shown in Fig.~\ref{fig:real-world}. These experiments demonstrate that FloCOS predictions can support collision-free planning in the tested physical interactions.

\subsection{Comparison of Computational Cost}
Table~\ref{tab:compute} compares model size and inference FLOPs across 32 development-set windows sampled at uniform index intervals (batch size 1, single target vehicle). The reported counts encompass all active subnetworks and frozen backbones: M2I includes its marginal, relation, and conditional branches, and DOS-Ellipse includes the MTR backbone plus its elliptical occupancy head.

\begin{table}[t]
\caption{Prediction Network Size and Computation}
\label{tab:compute}
\centering
\small
\begin{tabular*}{\columnwidth}{@{\extracolsep{\fill}}ccc@{}}
\toprule
Methods & Parameters ($10^6$) & FLOPs ($10^9$/window)\\
\midrule
MTR family$^\dagger$~\cite{shi2022motion} & 64.150 & 13.925\\
DenseTNT~\cite{gu2021densetnt} & 31.115 & 91.495\\
M2I~\cite{sun2022m2i} & 93.567 & 184.193\\
TrajFlow~\cite{yan2025trajflow} & 51.587 & 12.932\\
DOS-Ellipse~\cite{shao2024towards} & 64.341 & 13.933\\
\textbf{FloCOS} & \textbf{0.372} & \textbf{0.032}\\
\bottomrule
\end{tabular*}
\tabnote{$^\dagger$MTR, MTR-GMM + FORCE-OPT, and MTR-Top1 + MMCPR-Joint share the same network statistics. FLOPs are measured per planning window for a single target vehicle, excluding elementwise operations, preprocessing, set postprocessing, and planning.}
\end{table}

With \SI{0.372}{M} parameters and \SI{0.032}{GFLOPs} per window, FloCOS requires approximately two orders of magnitude fewer parameters and operations than the transformer-based baselines in Table~\ref{tab:compute}.

\section{Conclusion}
We presented FloCOS, a conditional flow-matching framework that predicts compact forward occupancy sets for vehicle motion planning. FloCOS maintains high occupancy coverage while reducing occupied area on WOMD, and achieves the highest interaction completion rate under a unified planner. Real-world experiments further show that using FloCOS predictions enables collision-free planning for ego vehicles. Future work will extend the method to other traffic participants and conduct larger-scale real-world evaluation.
\bibliographystyle{IEEEtran}
\bibliography{references}
\end{document}